\documentclass[letterpaper]{article} 
\usepackage{aaai24}  
\usepackage{times}  
\usepackage{helvet}  
\usepackage{courier}  
\usepackage[hyphens]{url}  
\usepackage{graphicx} 
\def\UrlFont{\rm}  
\usepackage{natbib}  
\usepackage{caption} 
\usepackage{algorithm}
\usepackage{algorithmic}
\usepackage{amsfonts}
\usepackage{amsmath}

\usepackage{newfloat}
\usepackage{listings}
\DeclareCaptionStyle{ruled}{labelfont=normalfont,labelsep=colon,strut=off} 
\floatstyle{ruled}
\newfloat{listing}{tb}{lst}{}
\floatname{listing}{Listing}
\title{Robust Visual Imitation Learning with Inverse Dynamics Representations}
\author{
    Siyuan Li\textsuperscript{\rm 1}\equalcontrib, Xun Wang\textsuperscript{\rm 2}\equalcontrib, Rongchang Zuo\textsuperscript{\rm 1}, Kewu Sun\textsuperscript{\rm 1}, Lingfei Cui\textsuperscript{\rm 2}, Jishiyu Ding\textsuperscript{\rm 2}, \\
    Peng Liu\textsuperscript{\rm 1}, Zhe Ma\textsuperscript{\rm 2} \\
    \textsuperscript{\rm 1}Harbin Institute of Technology
    \\
    \textsuperscript{\rm 2}Intelligent Science \& Technology Academy Limited of CASIC
}
\affiliations{

}

\usepackage{bibentry}

\begin{document}

\maketitle

\begin{abstract}
Imitation learning (IL) has achieved considerable success in solving complex sequential decision-making problems. However, current IL methods mainly assume that the environment for learning policies is the same as the environment for collecting expert datasets. Therefore, these methods may fail to work when there are slight differences between the learning and expert environments, especially for challenging problems with high-dimensional image observations. 
However, in real-world scenarios, it is rare to have the chance to collect expert trajectories precisely in the target learning environment.
To address this challenge, we propose a novel robust imitation learning approach, where we develop an inverse dynamics state representation learning objective to align the expert environment and the learning environment.
 With the abstract state representation, we design an effective reward function, which thoroughly measures the similarity between behavior data and expert data not only element-wise, but also from the trajectory level.
 We conduct extensive experiments to evaluate the proposed approach under various visual perturbations and in diverse visual control tasks. Our approach can achieve a near-expert performance in most environments, and significantly outperforms the state-of-the-art visual IL methods and robust IL methods.
\end{abstract}

\section{Introduction}

Imitation learning (IL) has gained encouraging success in various domains, e.g., games \citep{scheller2020sample, vpt}, robotics \citep{vrepl-example0, rn}, and autonomous driving \citep{vrepl-example1, modelbased0}. By learning behaviors directly from expert demonstrations, IL provides a way of sparing the burden of designing delicate reward functions \citep{il-survey, arora2021survey}.
However, every coin has two sides. Despite no dependence on the rewards, collecting expert datasets requires much effort. Furthermore, in some scenarios, it is quite difficult to collect expert trajectories exactly in the target learning environment. 
For example, although the aim is to learn a manipulation policy for a real robot arm, since the hardware is sophisticated and expensive, the expert demonstrations could only be collected in a simulator, which is similar to the target real-world environment, but has differences.  
The difference between the expert environment and the target learning environment induces an important challenge for current IL approaches, especially for tasks with high-dimensional visual observations.

\begin{figure}[t!]
\centering
\includegraphics[width=0.4\textwidth]{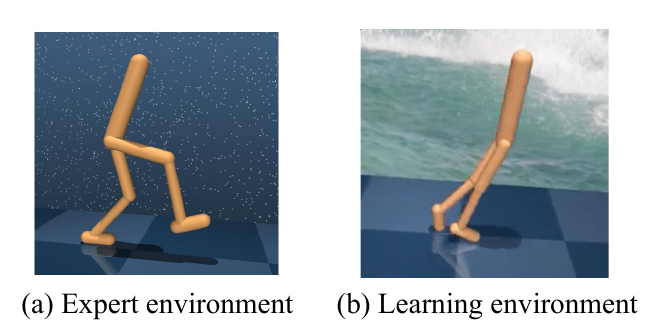}
\setlength{\belowcaptionskip}{-0.5cm}
\caption{We refer the environment of collecting the expert dataset as \textit{expert environment}, and the environment of learning the target policy as \textit{learning environment}.}
\label{fig0}
\end{figure}

\begin{figure*}[t!] 
\centering 
\includegraphics[width=0.9\textwidth]{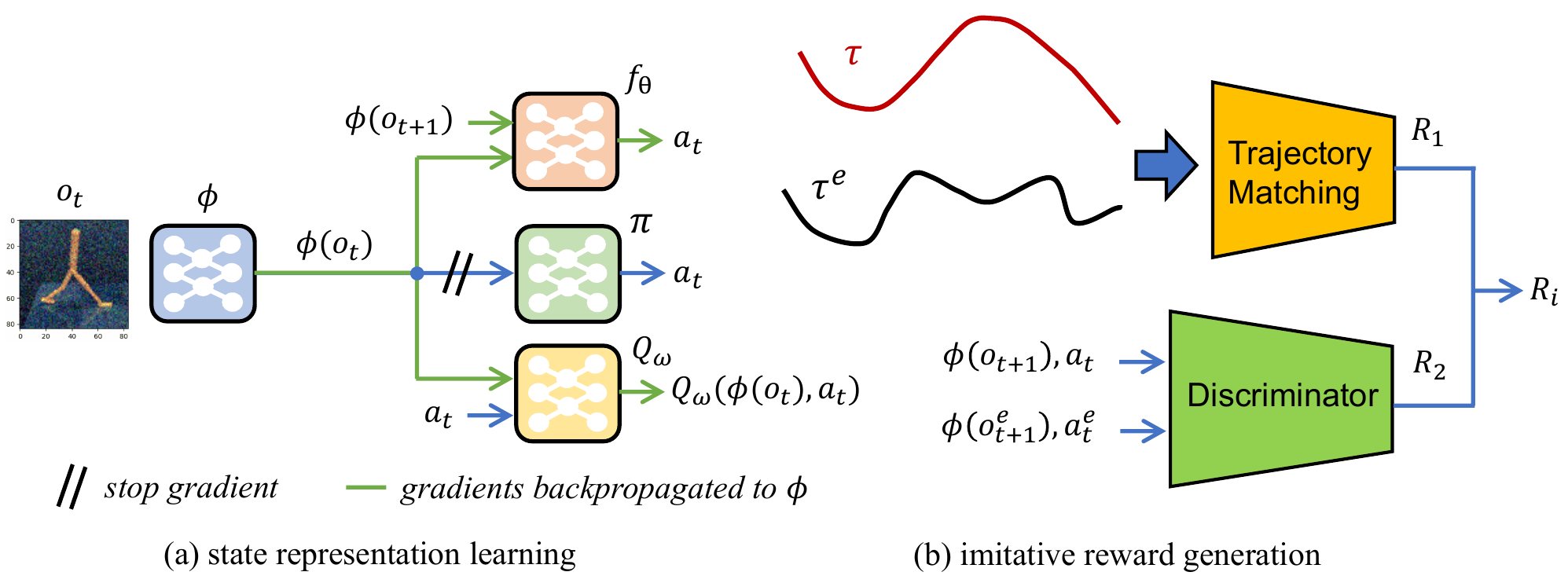} 
\caption{Robust visual imitation learning with inverse dynamics representations.} 
\label{Fig1} 
\end{figure*}

Behavior cloning (BC) \citep{bc0, bc1, bc2} is a conventional IL method, which maximizes the likelihood of taking the demonstrated action in a supervised learning manner. Although simple, BC requires a large amount of expert data and suffers from compounding error \citep{bc-shift0, bc-shift1}. As there is no online interaction, BC cannot handle the difference between the expert environment and the learning environment. 
Inverse reinforcement learning (IRL) \citep{irl0, irl1, irl2} is a popular learning paradigm for IL problems, which generates rewards by measuring the difference between expert data and behavior data. With the generated reward function, IRL employs an off-the-shelf RL algorithm to learn policies.
Note that in challenging tasks with high-dimensional observation space, estimating differences between observations is difficult. 
Furthermore, when the learning environment and the expert environment are not exactly the same, IRL approaches may misuse the difference between environments to generate uninformative rewards. For example, even if the robot positions in Figure \ref{fig0}(a) and Figure \ref{fig0}(b) are similar (both standing), 
the background in Figure \ref{fig0}(b) may disturb the reward generation, which leads to a small reward to punish the observation in Figure \ref{fig0}(b).  In fact, as the underlying state of this observation is similar to that of the expert observation, this observation should be encouraged. 
In addition, current IRL approaches measure the difference between expert data and behavior data either from the element-wise view \citep{ho2016generative, kostrikovdiscriminator, regularizeddiscriminator} or from the trajectory view \citep{haldar2023watch, papagiannis2023imitation, dadashi2021primal}, which does not fully utilize the expert dataset to generate effective rewards.

To align the learning environment and the expert environment, we propose a novel imitation learning approach based on inverse dynamics representation learning. 
The inverse dynamics objective serves to extract the action-related features from the high-dimensional observations, so that we could obtain a common state representation space between the learning and expert environments. 
Thanks to the inverse dynamics state representation, the proposed approach could deal with challenging IL tasks with visual observations, and is robust to the difference between the learning environment and the expert environment. 
By measuring the similarity between expert state embeddings and behavior state embeddings, we develop an imitative reward function, which not only considers the element-wise similarity of observation-action pairs, but also takes the trajectory-level similarity into consideration. 
This thorough reward function improves the previous IRL methods which only generate rewards from a single perspective.

We conduct extensive experiments on a set of visual control tasks in Meta-World domain \citep{yu2020meta} and DeepMind Control Suite (DMC) \citep{tassa2018deepmind}. The experiment results demonstrate that the proposed approach significantly outperforms the state-of-the-art visual IL methods and robust IL methods in terms of learning efficiency and convergent performance. To probe into the reason for the great performance, we further analyze the learned state representation in detail. Moreover, we conduct several ablation studies to validate the effectiveness of the various components in the proposed approach. 
It is hoped that these results could provide some insights for representation learning and reward design in robust visual imitation learning.

\section{Preliminaries}
We formulate the learning problem with a discounted finite-horizon Markov Decision Process (MDP). The MDP is of the form $(O, A, P, R, \gamma)$, where $O$ is the observation space, $A$ is the action space, $P(o_{t+1}|o_t, a_t)$ is the transition function specifying the probability distribution over the next state given the current state and action, $R: O\times A \rightarrow \mathbb R$ is the reward function, and $\gamma \in [0, 1)$ is the discount factor. In this paper, we focus on the challenging problems where the observations are high-dimensional images. The goal is to learn a policy $\pi: O \rightarrow A$ that maximizes the expected cumulative discounted reward: $\max_\pi\mathbb E_{P,\pi}[\sum_{t=1}^T\gamma^t R(o_t, a_t)]$, where $T$ denotes the horizon length.

In the IL setting, there is no available reward function for an agent to infer. Instead, the agent is provided with a demonstration dataset $\mathcal{T}^e = \{\tau_n^e|_{n=1}^N\}=\{(o_t^e, a_t^e)_{t=1}^T|_{n=1}^N\}$, which includes $n$ trajectories collected by experts. 
IRL approaches \citep{irl0,abbeel2004apprenticeship} try to solve the IL problem by generating rewards based on expert trajectories $\mathcal{T}^e$, and then optimize policy $\pi$ to maximize the cumulative rewards.
Our work falls in the IRL paradigm, and uses the actor-critic learning framework to conduct policy optimization.
Specifically, we employ the Twin Delayed Deep Deterministic policy gradient algorithm \citep{fujimoto2018addressing} as the base RL optimizer, which alleviates the Q overestimation issue with the twin delayed critic networks.

\section{Approach}
\label{approach}
This paper considers an IL setting where the learning environment is different from the environment of collecting the expert dataset (expert environment), e.g., the learning environment is perturbed by visual distractors (e.g., the white noise in $o_t$ in Figure \ref{Fig1}), but the expert environment is free from noise. The similarity between the learning and expert environments is that they share the same task. Previous IRL methods \citep{ho2016generative, torabi2018generative, kostrikovdiscriminator} generate rewards based on the similarity between expert data and behavior data. However, under the visual perturbations, it is challenging to accurately measure the similarity of the underlying states, which severely hurts imitation policy learning.

To fully exploit the expert trajectories even not in the expert environment, we propose a Robust visual Imitation Learning approach based on Inverse dynamics Representations (RILIR), as depicted in Figure \ref{Fig1}. To avoid the negative influence of visual perturbations, we design a state representation module to extract prominent features from image observations.
 Based on the abstract state representation, we design a reward function that measures the similarity between the learned policy and the expert policy from not only the single-step perspective but also the trajectory perspective. The following of this section first elaborates on the state representation learning module, and then describes the way of using the state embeddings to generate rewards. 

\subsection{State Representation Learning}
\label{sec31}
To facilitate reward generation and imitation policy learning, we aim to learn a representation function $\phi$, which extracts  state embedding $z_t = \phi(o_t)$ from high-dimensional observation $o_t$. 
The features related to actions are important for decision-making, and with the action-related state representation, we could better measure the similarity between the behavior trajectories and the expert trajectories.
To learn such state representations, we build an inverse dynamics network $f_\theta(\phi(o_t), \phi(o_{t+1}))$, which takes the subsequent state embeddings as inputs and predicts the action $a_t$ in the transition. 
In addition to the action-related property, the state representation needs to be beneficial to imitation policy learning as well, since state representation learning and policy learning influence each other in a loop: the training data for the state representation is collected by the learned policy, and the rewards for policy learning are dependent on the learned state representation.
Therefore, we augment the inverse dynamics objective with the value function optimization in an end-to-end manner.
 The overall loss function of representation network $\phi$, inverse dynamics network $f_\theta$, and the Q network $Q_\omega(\phi(o_t), a_t)$ is as follows:
\begin{equation}
\begin{aligned}
\label{eq1}
&L_{\phi, \theta, \omega} = \mathbb E_{(o_t, a_t, o_{t+1})\sim \{ \tau, \tau^e\}}[(f_\theta(\phi(o_t), \phi(o_{t+1}))-\hat{a}_t)^2] \\
& +\mathbb E_{(o_t, a_t, o_{t+1})\sim \tau, r_t=R_{i}(\tau, \tau^e)} [(r_t+\gamma \min_{i=1, 2}\hat{Q}^i(\phi(o_{t+1}), a_{t+1})\\
&\quad \quad \quad -Q_\omega^i(\phi(o_t), a_t))^2|_{i=1,2}],
\end{aligned}
\end{equation}
where $R_{i}$ denotes the imitative reward function, $\hat{Q}$ denotes the target Q network, and $a_{t+1}$ is the action taken by the policy $\pi$ on observation $o_{t+1}$ with exploration:
\begin{equation}
a_{t+1}\leftarrow \pi(\phi(o_{t+1}))+\epsilon, \epsilon \sim clip(\mathcal{N}(0, \sigma), -c, c).
\end{equation}
The loss in Equation \eqref{eq1} influences the representation network $\phi$ through stochastic gradient descent, as depicted by the reverse directions of the green lines in Figure \ref{Fig1}(a).

To boost the diversity of the training data for state representation, the inverse dynamics objective is optimized with both the expert and behavior trajectories. 
As we focus on the tasks with continuous action space, the inverse dynamics network is optimized with the mean-squared loss. For tasks with discrete action space, the first term in Equation \eqref{eq1} could be alternated with a cross-entropy loss.
 Note that the online update of the representation function induces non-stationarity for the rewards, we employ a target representation network for reward generation, where the target representation network is synchronized with the learned representation network for each $\Delta t$ timesteps.  

\subsection{Imitative Reward Generation}

 A key component in IRL is generating effective rewards to learn policies resembling the expert policy. The rewards generated by the discriminator in the adversarial IRL frameworks suffer from the non-stationary issue, as the discriminator is updated online with policy learning. Recent IL works \citep{cohen2021imitation,dadashi2021primal,haldar2023watch,papagiannis2023imitation} propose to generate rewards with trajectory matching, which measures the similarity between the expert trajectories and the behavior trajectories with optimal transport. Since trajectory matching is non-parameteric, the rewards in these methods are stationary. However, as the optimal-transport based trajectory matching methods emphasize the trajectory similarity as a whole, the element-wised similarity, i.e. the similarity of state-action pairs, may be neglected. To solve this problem, we propose a novel reward function that combines the advantages of trajectory matching and state-action pair similarity. 
 The following of this subsection first gives the formulations of trajectory matching rewards and discriminator rewards, and then elaborates on how to integrate them.

 \textbf{Trajectory Matching Rewards} Inspired by previous works \citep{haldar2023watch}, we compute the closeness between the expert trajectories $\mathcal{T}^e$ and behavior trajectories $\mathcal{T}$ by measuring the optimal transport of probability mass from $\mathcal{T} \rightarrow \mathcal{T}^e$. Given a cost function $c:\mathcal{Z}\times \mathcal{Z} \rightarrow \mathbb{R}$ defined in the state representation space $\mathcal{Z}$ and an optimal transport objective $g$, the optimal alignment between an expert trajectory $\tau^e$ and a behavior trajectory $\tau$ is shown in Equation \eqref{star}.\footnote{Here we overwrite the notation $f_\theta$ for trajectory embeddings: $f_\theta(\tau)=[f_\theta(o_1), ..., f_\theta(o_T)]$.} Note that it is necessary to define the cost function in the state representation space, since the distance in the high-dimensional image observation space cannot measure the similarity between states.
 \begin{equation}
 \label{star}
 \mu* \in \mathop{ \arg \min}\limits_{\mu \in \mathcal{M}} g(\mu, f_\theta(\tau), f_\theta(\tau^e), c),
 \end{equation}
where $\mathcal{M}=\{\mu \in \mathbb{R}^{T\times T}: \mu\mathbf{1}=\mu^\mathrm{T}\mathbf{1}=\frac{1}{T}\mathbf{1}\}$ is the coupling matrix set and the cost function $c$ could be Euclidean or cosine distance.\footnote{We provide an ablation study on the cost function $c$ in Appendix D, and using cosine distance performs slightly better than Euclidean distance.} Specifically, we utilize the Wasserstein distance with cosine cost as the optimal transport metric, and then
\begin{equation}
\begin{aligned}
g(\mu, f_\theta(\tau), f_\theta(\tau^e), c) &= \mathcal{W}^2(f_\theta(\tau), f_\theta(\tau^e)) \\
& =\sum_{t, t'=1}^T C_{t, t'}\mu_{t,t'},
\end{aligned}
\end{equation}
where the cost matrix $C_{t, t'} = c(f_\theta(o_t), f_\theta(o_{t'}^e))$. By maximizing the rewards in Equation \eqref{R1}, the agent could learn the policy that closely matches the expert trajectories,
\begin{equation}
\label{R1}
R_1(o_t)=-\sum_{t'=1}^TC_{t,t'}\mu^*_{t,t'}.
\end{equation}
Solving Equation \eqref{star} to compute $\mu^*$ is computationally expensive, so we employ the Sinkhorn algorithm \citep{knight2008sinkhorn} to obtain an approximate solution. As there are multiple expert trajectories, previous works \citep{haldar2023watch, cohen2021imitation} search the expert dataset, and use the nearest trajectory with the current behavior trajectory to compute the rewards in Equation \eqref{R1}. 
However, the ignored expert trajectories may contain certain state-action pairs, which are closer to the current behavior trajectories but not used to generate rewards, since a trajectory is considered as a whole. 
To thoroughly utilize the expert dataset, we propose to augment the trajectory-matching rewards with the following discriminator rewards, which estimate the similarity of state-action pairs.

 \textbf{Discriminator Rewards} Following the GAIL method \citep{ho2016generative}, we train a discriminator $D$ that differentiates between the agent's samples and the expert data with the following loss function:
 \begin{equation}
 \begin{aligned}
L_D = &- \mathbb E_{(o_t^e, a_t^e)\sim \mathcal T^e}[\log D(f_\theta(o_t^e), a_t^e)] \\
&- \mathbb{E}_{(o_t, a_t)\sim \mathcal T}[\log (1-D(f_\theta(o_t), a_t))].
\end{aligned}
 \end{equation}
 Similar to trajectory matching, the discriminator $D$ also works in the state representation space. Beyond that, $D$ takes the actions into account, hence the similarity between the expert trajectories and the behavior trajectories could be measured more accurately. By training the discriminator, we derive another reward function:
 \begin{equation}
R_2(o_t, a_t) = -\log D (f_\theta(o_t), a_t). 
 \end{equation}

To summarize, the trajectory matching reward is derived from a macro view, and the discriminator reward is formulated from a micro view. They are both heavily dependent on the state representation learning in Section \ref{sec31}.
By integrating these two kinds of rewards together, we obtain a reward function $R_i$ which could thoroughly describe the similarity between the expert data and the behavior data:
\begin{equation}
R_i(o_t, a_t) = R_1(o_t) + \eta R_2(o_t, a_t).
\end{equation}
$\eta$ is a scaling factor balancing these two kinds of imitative rewards. 
In Appendix A, we provide the pseudocode and the algorithmic details of RILIR.

\section{Related Work}
\label{related}
\textbf{Imitation Learning} IL aims to learn policies from demonstrations without access to the environment rewards \citep{il-survey}. There are three major paradigms in IL. (1) Behavior cloning (BC)~\citep{bc0,bc1,bc2} treats policy learning as a supervised learning problem over state-action pairs. While these methods are appealingly simple, they suffer from compounding errors caused by covariate shift~\citep{bc-shift0,bc-shift1}.
An improved BC method~\citep{bc-modify} alleviates the covariate shift problem in specific tasks. 
(2) Inverse reinforcement learning (IRL)~\citep{irl0,irl1,irl2} infers rewards from the given demonstrations.
Compounding error is not an issue for these methods \citep{irl-better0,irl-better1}. However, IRL is extremely expensive regarding samples, since after inferring rewards, it still needs to run RL methods in an inner loop to learn the policies.
(3) Generative adversarial imitation learning (GAIL)~\citep{ho2016generative} is an adversarial learning based formulation inspired by maximum entropy IRL~\citep{irl1} and GANs~\citep{gan}.  Compared to IRL, this line of research does not need to infer the rewards while regarding the discriminator results as an auxiliary \citep{dadashi2021primal,papagiannis2023imitation}, which contributes to more efficient learning. 
Previous works \citep{baram2017end, sun2021adversarial} combine model-based learning with GAIL to construct a fully differentiable frame and enable more accurate gradient estimation.
Different from these model-based IL methods, the inverse dynamics model in this work is designed for an abstract state representation, and the discriminator reward is inspired by the GAIL paradigm.

\textbf{Robust Imitation Learning} Robustness against the variations between learning and expert environments has recently received much attention. Domain adaptive IL methods \citep{il-robustness1, kim2020domain} seek the consistency between the expert and learning domains with a set of prior data pre-collected in both two domains, and use this consistency for policy learning in a related target task in the learning domain. In contrast to domain-adaptive IL, the proposed approach does not need a prior dataset, and can learn directly in the learning environment instead. \citet{il-robustness0} proposed an IL method to deal with the dynamics variance in the learning and expert environments. The proposed approach aims to solve the visual variance problem, which commonly occurs in real-world tasks, e.g., an occluded camera.
The SeMAIL method \citep{wan2023semail} tries to solve the visual distractors in IL with model-based learning. However, as the high-dimensional visual observations are hard to reconstruct, the forward model learning in SeMAIL is sample inefficient. In contrast, the proposed approach learns an inverse dynamics model, which has achieved a better performance, as shown in the experiment section.

\begin{figure*}[t!] 
\centering 
\includegraphics[width=1.0\textwidth]{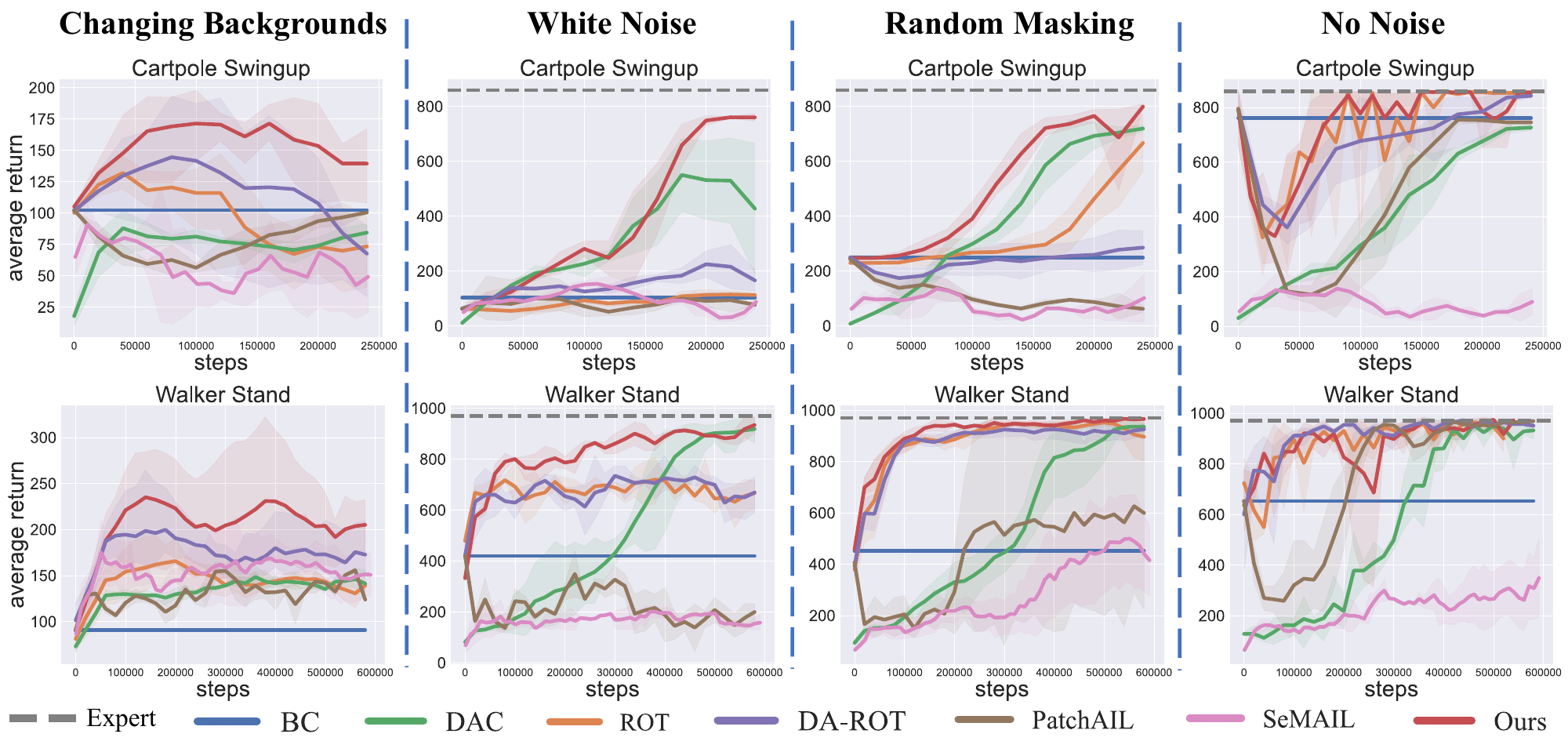} 
\setlength{\abovecaptionskip}{-0.2cm}
\setlength{\belowcaptionskip}{-0.5cm}
\caption{Experiment results under various types of visual shifts. The videos of the learned policies are shown in \protect\url{https://sites.google.com/view/rilir-aaai.}} 
\label{fig:new} 
\end{figure*}

\textbf{Visual Representation Learning (Visual RepL)} Learning from visual observations is an important problem due to its potential impact on fields like robotics~\citep{vrepl-example0}, autonomous driving~\citep{vrepl-example1} and video games~
\citep{vrepl-example2}.  OpenAI seeks to train general-purpose foundation models for sequential decision-making by utilizing freely available internet-scale unlabeled visual datasets via imitation learning~\citep{vpt}. However, this paradigm suffers from low sample efficiency and poor generalization ability due to the high-dimensional visual space. Therefore, Visual RepL plays a critical role, which could be divided into the following categories. 
(1) Representation learning with auxiliary objectives, e.g., learning compact state representations with an encoder-decoder architecture~\citep{encoderdecoder}, enhancing discriminators with patches~\citep{regularizeddiscriminator}, joint optimization of model, representation and policy~\citep{jointauxiliary}. 
(2) World model based methods~\citep{modelbased0,modelbased1, masked}. These methods first construct an evolution world by a latent dynamics model that predicts latent states from visual observations, and then the learned latent state is fed to the policy network as an input.
Rolling out within the learned world model can help reduce the sample number in the real environment for RL methods \citep{rlmodelbased0,rlmodelbased1}.  
(3) Image augmentation \citep{yaratsimage, curl}, which is widely studied in the computer vision domain~\citep{dataaugmentation0,dataaugmentation1}. Recently it has also been utilized to promote representation learning in RL and IL \citep{rn, empirical}.
Our work belongs to the auxiliary-task paradigm, and could be easily combined with the data augmentation methods.

\section{Experiments}
\label{exp}
We evaluate the proposed approach RILIR on a set of challenging visual control environments with perturbations, aiming at answering the following questions: 
(1) Can RILIR generally work under various types of visual perturbations?
(2) Can RILIR work in different types of tasks, including locomotion and manipulation?
(3) How is the state representation learned by RILIR? 
(4) How important are the various components of the RILIR approach?

\subsection{Experiment Results Under Various Visual Shifts}
\label{sec52}
In this subsection, we aim to evaluate the ability of RILIR to work under different types of visual shifts, and two tasks from the DeepMind Control (DMC) suite \citep{tassa2018deepmind} are used for evaluation.
RILIR's ability to work in diverse tasks is evaluated in the next subsection.
 RILIR is compared with state-of-the-art IL methods, including BC, adversarial IL methods, visual IL methods, and robust IL methods.
A brief description of the baselines is as follows, and in Appendix C, we provide the hyperparameters for all the baselines and the proposed approach.
\begin{itemize}
    \item Behavior cloning (BC): Supervised learning method trained with expert demonstrations.
    \item Discriminator Actor-Critic (DAC) \citep{kostrikovdiscriminator}: An adversarial IL method, which outperforms prior works such as GAIL \citep{ho2016generative} and AIRL \citep{fulearning}. 
    \item Regularized Optimal Transport (ROT) \citep{haldar2023watch}: A state-of-the-art visual IL method based on trajectory matching, which not only takes the BC policy as an initialization, but also regularizes the policy updates with the BC objective.
    \item DA-ROT \citep{yaratsimage}: As data augmentation has shown its strength in handling visual shifts, we compare RILIR with the DrQ data augmentation version of ROT.
    \item PatchAIL \citep{regularizeddiscriminator}: A visual IL method which generates rewards based on patches of images.
    \item SeMAIL \cite{wan2023semail}: A robust IL method which aims to eliminate visual distractors via separated models.
\end{itemize}

\begin{figure*}[t!] 
\centering 
\includegraphics[width=1.0\textwidth]{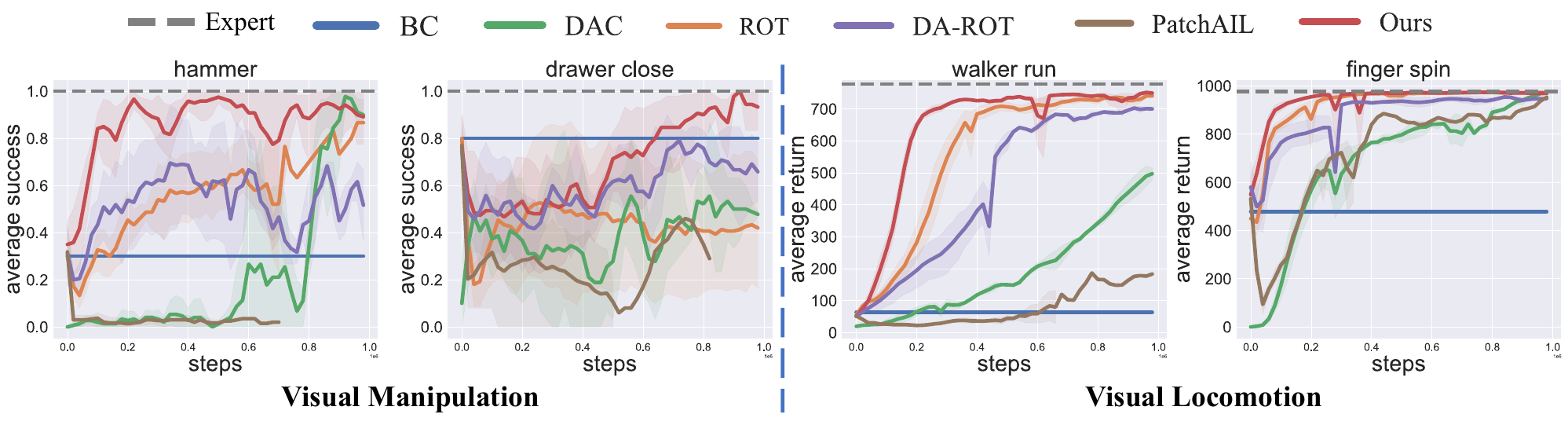} 
\setlength{\abovecaptionskip}{-0.2cm}
\setlength{\belowcaptionskip}{-0.5cm}
\caption{Experiment results in various types of tasks with random masking noises.} 
\label{fig3} 
\end{figure*}
In the experiments, we consider three types of visual shifts, including changing the backgrounds, adding white noise, and random masking. For the experiments of changing backgrounds, we use the environments in the easy version of the distracting control suite \citep{stone2021distracting}. 
In contrast to the noisy learning environments, the environments for collecting the expert demonstrations are free of noise. We provide the details of the learning environments and the way of collecting expert datasets in Appendix B. 
Besides, we conduct comparative experiments in ``clean'' environments without visual distractors as well. 
Figure \ref{fig:new} demonstrates the results in the CartPole Swingup task and the Walker Stand task under various types of visual shifts. The $y$ axis shows the average return over $10$ episodes. Each line is the mean of $3$ runs with shaded regions corresponding to a confidence interval of $95\%$. All the curves have been smoothed equally for visual clarity. The code to reproduce these results is available in the supplementary material.

As shown in the first three columns in Figure \ref{fig:new}, the proposed approach significantly outperforms the baseline methods.\footnote{These experiments have been run with A100 GPUs, and each run takes no more than 1 day.} In the ``clean'' environments, the performance of all the methods is nearly the same. 
Similar to ROT, our approach and PatchAIL also utilize the behavior-cloned policy as an initialization for the policy network, so there is a jump start in these learning curves.
Comparing the columns in Figure \ref{fig:new}, we find that changing backgrounds is severely more challenging than other visual shifts. Even in this challenging environment, the proposed approach can achieve positive learning and perform better than the baselines. In other types of environments, the proposed approach accomplishes a near-expert learning performance using much fewer samples than the baselines.

Note that ROT \citep{haldar2023watch} is a state-of-the-art visual IL method, which has shown better performance than DAC, and we also compare the proposed approach with the DrQ data augmented version of ROT (DA-ROT). 
Data augmentation improves the learning performance of ROT in some cases, but not all cases,
possibly because data augmentation induces a heavier computation burden and cannot cover all types of visual shifts.
PatchAIL has a similar performance with DAC in the ``clean'' environments, but the performance of PatchAIL drops severely in the noisy environments. This phenomenon implies that segmenting images into patches and generating rewards by taking averages over patches can hardly help the agent adapt to visual shifts. 
SeMAIL is a robust visual IL method with separated forward models. This method achieves a comparable performance with other baselines in the Walker Stand task with changing backgrounds.
However, SeMAIL is not as efficient as other baselines in the environments with easier visual perturbations due to the heavy burden of optimizing a forward model.
We provide the results of running SeMAIL with more steps in Appendix D.

\subsection{Experiment Results in Diverse Tasks}
To evaluate RILIR's ability to work in diverse tasks, we conduct experiments in two domains: manipulation and locomotion.  
Two tasks in the Meta-World benchmark \citep{yu2020meta} are used as the manipulation tasks, and two tasks in the DMC suite are used as the locomotion tasks.
These tasks are with medium difficulty, and the difficulty of tasks is measured by the dimension of action spaces. The learning results in harder tasks are provided in Appendix D.
To evaluate the robustness of these IL methods, the visual observations in the learning environments are randomly masked. 

As shown in Figure \ref{fig3}, in both manipulation and locomotion domains, the proposed approach significantly outperforms the baselines regarding learning efficiency and convergent performance.
Benefiting from the inverse dynamics representation learning and effective imitative rewards, the return or success gap between the policy learned by the proposed method and the expert policy is less than $5\%$.
Similar to the results in Figure \ref{fig:new}, PatchAIL suffer from the random masking noises.
In the following subsection, we analyze the state representation learning in our method to probe into the reason why it can achieve such a good performance.  

\subsection{Analysis of the State Representation}
In this subsection, we analyze the learned representation in the RILIR approach. 
Specifically, we visualize the state representations in the form of saliency maps \citep{simonyan2013deep} to analyze which regions have been paid more attention by the representation function. 
The saliency map is calculated as follows,
\begin{equation}
\sum_i|\frac{\partial(\phi_i(o_t))}{\partial o_t}|,
\end{equation}
where $\phi_i(o_t)$ denotes the $i$-th element of the state representation $\phi(o_t)$. A pixel with a larger saliency value has a larger influence on the Q network and the policy network, since $\phi(o_t)$ is the input of these two networks.

\begin{figure*}[htbp] 
\centering 
\includegraphics[width=0.8\textwidth]{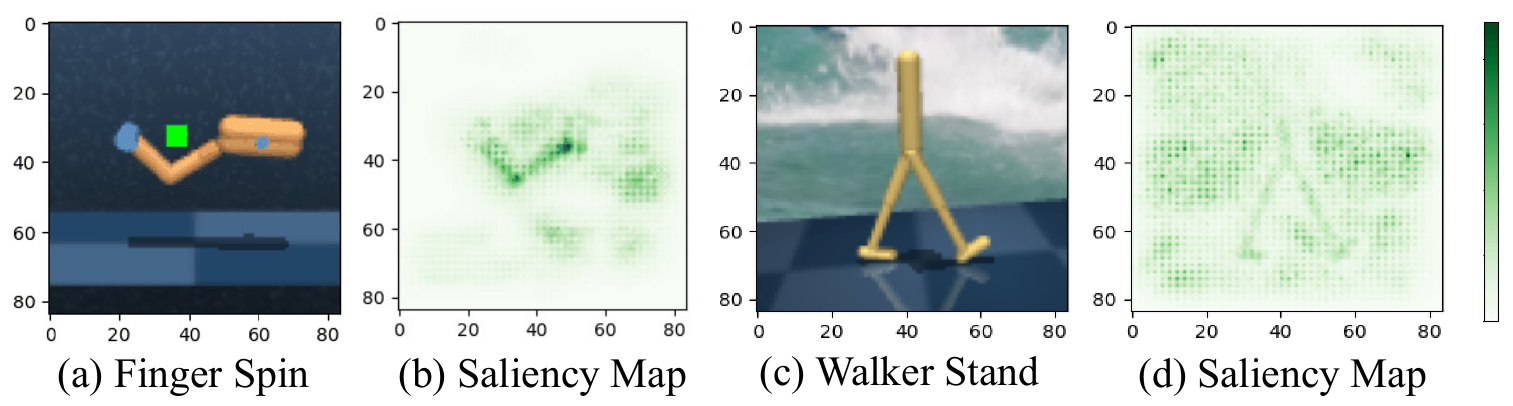} 
\caption{ (a)(b) Example observation in the Finger Spin task with random masking perturbation, and the corresponding saliency map at 200k steps.
(c)(d) Example observation in the Walker Stand task with changing background distractors from the Distracting Control suite \citep{stone2021distracting}, and the corresponding saliency map at 600k steps.
} 
\label{fig4} 
\end{figure*}

A darker green color denotes a larger saliency value. 
In the locomotion tasks, the proprioception features of the robot are substantially important.
As shown in Figure \ref{fig4}, our method clearly extracts the position of the \textit{Finger} in the Finger Spin task and ignores the random masking distractors. 
With a changing background, our approach can still extract the leg positions of the simulated Walker robot, as shown in Figure \ref{fig4} (c) and (d).
The state representations in more tasks are shown in \url{https://sites.google.com/view/rilir-aaai}, and we provide a comparison with the representations learned by the baselines in Appendix D.

\subsection{Ablation Studies}

\begin{figure}[htbp] 
\centering 
\includegraphics[width=0.48\textwidth]{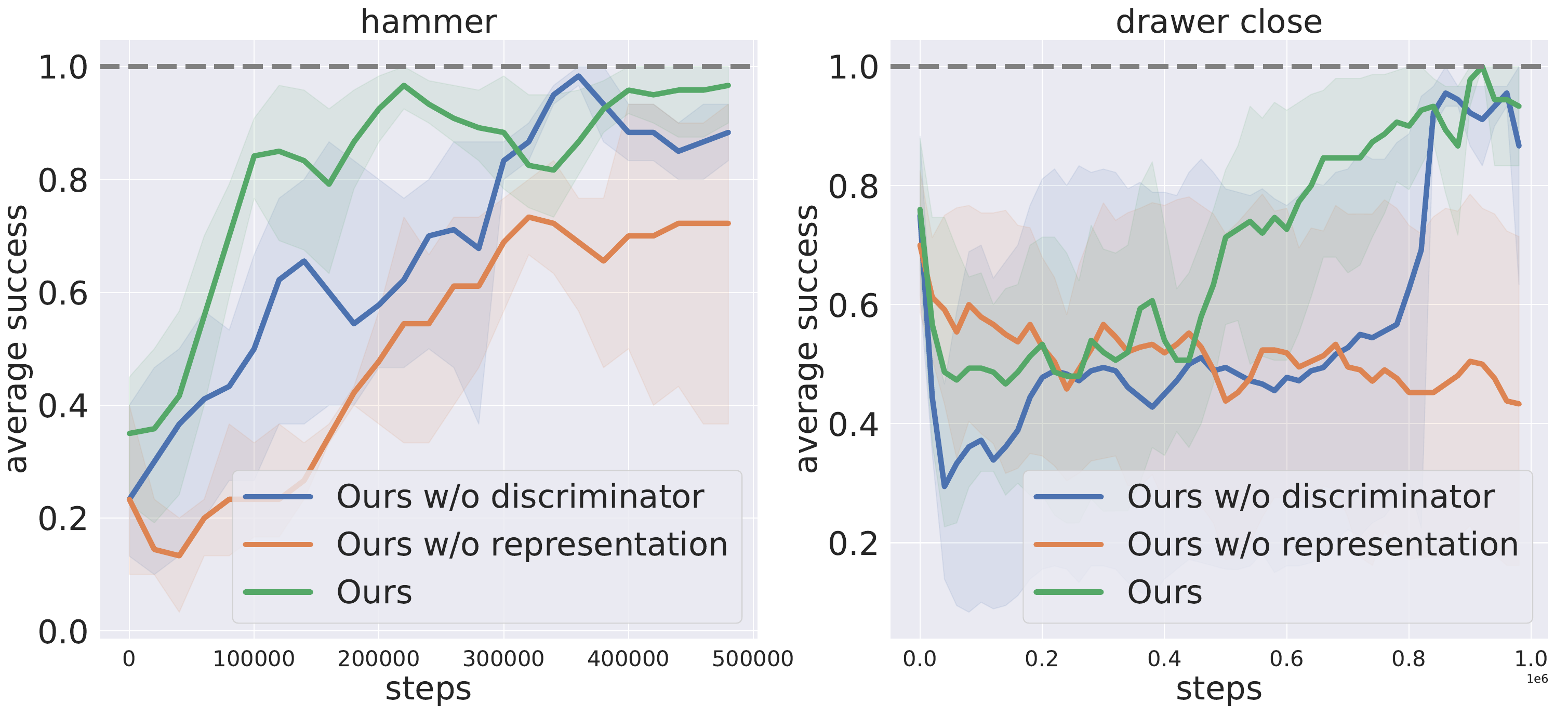} 
\caption{ Ablation studies of the state representation learning and the reward function in the proposed approach RILIR.
} 
\label{fig6} 
\end{figure}

To validate the effectiveness of various components in the proposed approach, we conduct ablation studies in two tasks, Hammer and Drawer Close, with random masking perturbations in the environments. Hammer is relatively easy, and Drawer Close is a difficult task, as it takes nearly $1$ million steps for RILIR to converge. In this subsection, we have respectively ablated the representation learning module and the rewards in RILIR. ``Ours w/o representation'' in Figure \ref{fig6} denotes the experiments removing the inverse dynamics representation objective, and ``Ours w/o discriminator'' denotes removing the discriminator rewards, i.e., only using the trajectory matching rewards.
More ablation studies in both the noisy environment and the ``clean'' environment are provided in Appendix D.

The results in Figure \ref{fig6} show that both the representation learning module and the discriminator rewards are effective. Specifically, the discriminator rewards mainly affect learning efficiency, and the inverse dynamics representation learning influences convergent performance. Beyond that, in more challenging tasks (Drawer Close), the influence is more obvious.
Furthermore, among the three curves, the shaded areas of the proposed approach are the smallest. This implies that the state representation learning objective and the compound rewards help stabilize the learning process. 

\section{Conclusion and Limitations}
In this work, we have proposed a robust visual imitation learning approach based on inverse dynamics representation learning, dubbed RILIR, which is able to resist the difference in the expert environment and the learning environment, since the representation module extracts the common parts in these two environments. Based on this abstract state representation, we develop a thorough reward function, which considers the similarity of expert data and behavior data from both an element-wise view and a trajectory-level view.
Extensive experiment results in a set of challenging visual control tasks demonstrate that the proposed approach has achieved substantially better performance than prior visual IL works and robust IL works. 
Furthermore, we have conducted ablation studies to validate the effectiveness of the representation learning module and the reward function in the RILIR approach. 

However, we recognize a few limitations in this work: 
(a) The proposed approach may not work well when the observations in the learning environments are significantly different from those in the expert demonstrations, and more advanced representation learning methods need to be investigated in the robust visual IL domain. 
(b) The inverse dynamics objective may not be the right thing when the dynamics models are not consistent between the learning environment and the expert environment, as this objective relies on the invariant dynamics model. 
A recent work \citep{il-robustness0} proposed to solve the robust IL problem with various dynamics by imitating multiple experts simultaneously.


\end{document}